# OVERVIEW OF THE ARABIC SENTIMENT ANALYSIS 2021 COMPETITION AT KAUST

**Hind Alamro**
KAUST
Saudi Arabia

**Manal Alshehri**
KAUST
Saudi Arabia

**Basma Alharbi**
University of Jeddah
Saudi Arabia

**Zuhair Khayyat**
Lucidya
Saudi Arabia

**Manal Kalkatawi**
KAU
Saudi Arabia

**Inji Ibrahim Jaber**
KAUST
Saudi Arabia

**Xiangliang Zhang** *
KAUST
Saudi Arabia

## ABSTRACT

This paper provides an overview of the Arabic Sentiment Analysis Challenge organized by King Abdullah University of Science and Technology (KAUST). The task in this challenge is to develop machine learning models to classify a given tweet into one of the three categories *Positive*, *Negative*, or *Neutral*. From our recently released ASAD dataset, we provide the competitors with 55K tweets for training, and 20K tweets for validation, based on which the performance of participating teams are ranked on a leaderboard, https://www.kaggle.com/c/arabic-sentiment-analysis-2021-kaust. The competition received in total 1247 submissions from 74 teams (99 team members). The final winners are determined by another private set of 20K tweets that have the same distribution as the training and validation set. In this paper, we present the main findings in the competition and summarize the methods and tools used by the top ranked teams. The full dataset of 100K labeled tweets is also released for public usage, at https://www.kaggle.com/c/arabic-sentiment-analysis-2021-kaust/data.

## 1 Introduction

Sentiment Analysis (SA) is a widely studied problem in Natural Language Processing (NLP). It is the task of automatically detecting and identifying the sentiment of a written text, usually by labeling a piece of text as *Positive, Negative*, or *Neutral* sentiment. In recent years, the growth of social media allows people to share their own ideas, thoughts, and emotions in public. In Twitter, for instance, users generate huge amounts of text data containing their insights [1]. Consequently, Sentiment Analysis of tweets becomes popular and has been studied for different purposes, from analyzing sentiments in general to specific fields. The specific fields include marketing analysis of services or products [2, 3], political field analysis [4, 1, 5], and public action towards events, persons, and recently, pandemics [6, 7, 8].

Arabic is one of the most popular languages in the world. It is the official language for 27 countries located in the Middle East and North Africa. Recently, Arabic natural language processing has attracted a lot of attention due to the increase usage of the language over the web and social media. Arabic exists in different forms and dialects according to each country or area. The formal form of Arabic is used in formal purposes like education and news [9]. However, in social media like Twitter, the informal and the free writing forms of the language are widely used and make the studies of Arabic tweets more complicated [10].

We launch an "Arabic Sentiment Analysis Competition" for promoting the study in this topic [11]. The task in this competition is to analyze the sentiments of Arabic tweets by classifying a given tweet text into one of the three categories: *Positive, Negative*, or *Neutral*. The competition is sponsored by KAUST. The award to top-3 winners is 17000 USD in total. It runs on Kaggle[2], a popular platform for Machine Learning Competitions.

---

*The corresponding author: Xiangliang Zhang. Email: xiangliang.zhang@kaust.edu.sa

The competition is based on our released benchmark dataset ASAD [11], which is a large collection of 100K Arabic tweets, annotated for sentiment analysis tasks. We provided the competitors with training and testing files. The training file contains 55K annotated tweets, and the testing file (TEST1) contains 20K tweets, based on which the performance of participation teams is ranked on a leaderboard. Additionally, we kept another testing file (TEST2) of 20K tweets, which was not released to the competitors and used as a private testing file.

Seventy-four teams participated in this competition and in total made 1247 submissions. We summarize the techniques, methods, and tools used by the top-ranked teams in this paper. Basically, all the participants used some preprocessing techniques, such as removing the URLs, hashtags, user ids, foreign characters, and repeated characters. In general, all the top-ranked teams used the state-of-the-art (SOTA) pre-trained language model MARBERT to learn new representation of tweets. MARBERT [12] is a large-scale Arabic pre-trained language model that focuses on both dialectal Arabic and modern standard Arabic. However, each team has its own approach for how to use MARBERT and other different deep learning techniques. Section 3 describes these approaches in details.

For determining the winners, we followed two steps. First, we ranked the teams by their performance on TEST1, as shown in the leaderboard. Second, we invited the top-ranked teams to submit their codes for the evaluation on TEST2, which has the same distribution as TEST1 and the training dataset. The final winners were determined based on the results of TEST2. More details can be found in Section 2 of the competition introduction.

The rest of this paper is organized as follows. Section 2 describes the competition task, the dataset, the evaluation process and baseline models. Section 3 discusses the participants' results and the used models of the top three winning participants. Section 4 introduces other tasks with similar objective; i.e., Arabic sentiment analysis. Finally, Section 5 provides concluding remarks on KAUST 2021 Arabic sentiment analysis competition.

## 2 The Competition

The Arabic sentiment analysis competition is a multi-class classification task, where sentiment is identified at a 3-point scale. The goal is to classify the sentiment of a tweet as either: *positive, negative* or *neutral*. The competition was hosted on Kaggle framework and sponsored by KAUST. The prizes for the first three winners were 10000 USD, 5000 USD and 2000 USD, respectively.

### 2.1 Dataset Description

The competition is organized by using our ASDA dataset, which is the largest to this date for Arabic tweets sentiment analysis. It contains 100K tweets annotated as either *positive, negative* or *neutral*. The tweets in ASAD were collected in the period between May 2012 and April 2020. They are written in different Arabic dialects including Khaleeji, Hijazi, Egyptian and Modern Standard Arabic. Table 1 shows some examples of the annotated tweets from ASAD. The data collection and annotation were done by Lucidya[3] which is an AI-based company with rich experience in organizing data annotation projects. A detailed description of our ASAD dataset and the process of collection and annotation can be found in [11]. The distribution of the released data in competition can be found in Table 2.

The full ASDA dataset is available to the research community to be used freely beyond the competition. The dataset containing tweet IDs and annotations is available for free access[4]. Additionally, to simplify the process of obtaining tweets from tweet IDs, we provide a platform in which users can supply tweet IDs and get the actual tweet content in return. To get access to this platform, users can register in the competition's website[5] and freely use the online platform.

### 2.2 Participation Teams

At the time of writing, a total of 764 users registered in the competition's website, from 45 different countries. Figure 1a illustrates the number of registered users per country for the top five countries. We can see that most registered users came from Saudi Arabia, followed by Algeria then Egypt. Figure 1b visualizes the number of downloaded tweets, using the platform, per month. From the figure, we can conclude that the largest number of tweets were downloaded in January. Tweets continued to be downloaded using the platform through the months from February to June, 2021.

---

[2]https://www.kaggle.com/c/arabic-sentiment-analysis-2021-kaust
[3]https://lucidya.com/
[4]https://www.kaggle.com/c/arabic-sentiment-analysis-2021-kaust/data
[5]https://wti.kaust.edu.sa/solve/Arabic-Sentiment-Analysis-Challenge

Table 1: Example of tweets in ASAD

| Tweet | Sentiment |
|---|---|
| لأول مره أفرح بدرجه في الكليه القمر اللي انا فيها دي ف الحمد لله بجد ؟؟ | Positive |
| احس بان مصر ستكون فى الأيام القدمة اكثر اشراقا وأكثر قوة | Positive |
| أشعرُ مرةً أخرى برغبةٍ عارمة في التخلي عن كُل شيء. | Negative |
| هو انا ليه كدا عمال بكتشف في ناس وحششه | Negative |
| السلام عليكم ..كم يوم تأخذ نتيجة تحليل فايروس كورونا ؟ | Neutral |
| القوة ليست دائما فيما نقول ونفعل أحيانًا تكون فيما نصمت عنه ، فيما نتركه بإرادتنا، وفيما نتجاهله. | Neutral |

Table 2: Class distribution in data splits in competition. The *All* set represents the complete competition dataset. *TRAINING, TEST1* and *TEST2* are the three splits used in the competition and also in the benchmark models evaluation. The three splits maintained the same class distribution as the original complete dataset.

|  | TRAINING | | TEST1 | | TEST2 | | All | |
|---|---|---|---|---|---|---|---|---|
|  | No. tweets | (%) | No. tweets | (%) | No. tweets | (%) | No. tweets | (%) |
| **Positive** | 8821 | 0.16 | 3150 | 0.16 | 3244 | 0.16 | 15215 | 0.16 |
| **Negative** | 8820 | 0.16 | 3252 | 0.16 | 3195 | 0.16 | 15267 | 0.16 |
| **Neutral** | 37359 | 0.68 | 13598 | 0.68 | 13561 | 0.68 | 64518 | 0.68 |
| **Total** | 55000 | 1.00 | 20000 | 1.00 | 20000 | 1.00 | 95000 | 1.00 |

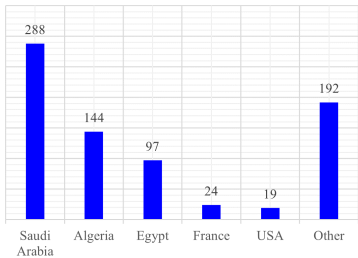

(a) Distribution of registered users per country, for the top five countries.

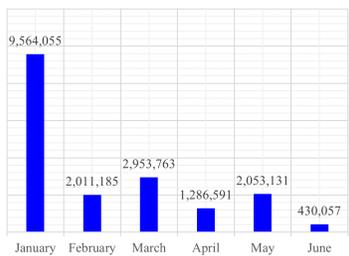

(b) Distribution of downloaded tweets per month, from January to June, 2021.

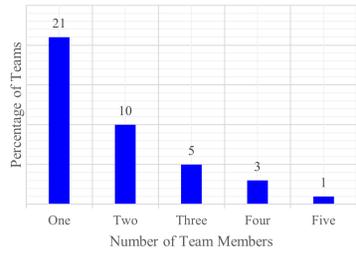

(c) Distribution of participation team sizes.

Figure 1: Statistics of registration and participation teams.

### 2.3 Evaluation Metrics and Baselines

For the multi-class sentiment analysis, we used the Macro-F1 score as the official evaluation metric. Evaluation was done in two phases: In phase I, we considered the results of TEST1 ranked by Macro-F1 in Kaggle's leaderboard. Then, in phase II, we invited the top-ranked teams to submit their codes for the evaluation on TEST2. With the submitted codes, we first checked the codes and verified the submitted results of TEST1. Then, we did the final evaluation on TEST2. The final winners were determined based on the Macro-F1 scores of TEST2.

We run several baseline models in [11]. The best model is based on AraBERT [13], which achieved an Macro-F1 score of 0.68 on both TEST1 and TEST2. We were expecting the winners to achieve much higher Macro-F1 scores than 0.68.

## 3 The Winning Teams

Seventy-four teams participated in this challenge, with a total of 1,247 submissions. The majority of teams (81%) consisted of one member only, and the remaining groups had 2-5 members. Figure 1c illustrates the distribution of team sizes for all participating teams. Table 3 lists the leaderboard from Kaggle, which ranks all teams by their Macro-F1 score on TEST1. Table 4 shows the performance of the top 20 teams on TEST1 in terms of accuracy, precision, recall, Micro-F1, Macro-F1, and weighted-F1 scores.

Table 3: Results achieved by all teams. Teams are ordered by the main evaluation metric, Macro-F1.

| ID | Team Name | Score | ID | Team Name | Score | ID | Team Name | Score |
|---|---|---|---|---|---|---|---|---|
| 1 | GOF | 0.79986 | 26 | DGGL | 0.73666 | 51 | Catherine Horbach | 0.65959 |
| 2 | Ahmed Elbehiry | 0.79979 | 27 | [Deleted] | 0.73601 | 52 | Hussein Ghaly | 0.65446 |
| 3 | Wissam Antoun | 0.7962 | 28 | Vision-AJ | 0.73337 | 53 | shimaa | 0.6511 |
| 4 | CS-UM6P | 0.79461 | 29 | Ibra | 0.72891 | 54 | aliashraflover14 | 0.64493 |
| 5 | Ali Salhi | 0.79349 | 30 | Alkhalifa | 0.72701 | 55 | Al_Salah | 0.64112 |
| 6 | Taicheng Guo | 0.79097 | 31 | Amal | 0.72642 | 56 | THH | 0.64029 |
| 7 | [Deleted] | 0.7909 | 32 | Vadim Yermakov | 0.72439 | 57 | Uliana Kobzar | 0.62815 |
| 8 | Salma Jamal | 0.7909 | 33 | Abdulshaheed Alqunber | 0.72294 | 58 | PFSA Team | 0.627 |
| 9 | Abdullah I. Alharbi | 0.79014 | 34 | faiz faiz | 0.72097 | 59 | OMA | 0.62411 |
| 10 | Aggies | 0.78753 | 35 | Jana Muhammad | 0.71852 | 60 | Egor Gavrilenko | 0.61251 |
| 11 | KUIS AI | 0.78735 | 36 | Abdullah Hussein | 0.71503 | 61 | laouni MAHMOUDI | 0.60329 |
| 12 | AraBrain Hidden Layers | 0.78482 | 37 | adamHesham | 0.71097 | 62 | dream | 0.58882 |
| 13 | AEM | 0.77837 | 38 | Enigma | 0.70668 | 63 | Shahad Althobaiti | 0.58322 |
| 14 | Omar Mohamed | 0.77455 | 39 | Horizon | 0.70238 | 64 | Mohammed Salem | 0.57911 |
| 15 | EAM | 0.77248 | 40 | Mikita Daroshkin | 0.69881 | 65 | Mazin Taha | 0.5753 |
| 16 | NLP players | 0.77074 | 41 | Yousef Nafea | 0.69564 | 66 | Nour samia | 0.50982 |
| 17 | raghad | 0.7691 | 42 | cher zola | 0.69491 | 67 | Hassan | 0.32578 |
| 18 | Murtadha Aljubran | 0.76704 | 43 | nemo | 0.68915 | 68 | ArabicNLP | 0.32353 |
| 19 | Roobaea Alroobaea | 0.7605 | 44 | Vladimir | 0.68854 | 69 | Khaled Al-Shamaa | 0.31951 |
| 20 | Yolo | 0.75998 | 45 | François de Ryckel | 0.68738 | 70 | Abdulelah | 0.30239 |
| 21 | Marwa Gharbi | 0.75917 | 46 | Manal Mohammed | 0.68731 | 71 | Yuchen Li | 0.29423 |
| 22 | Husain Khatba | 0.7548 | 47 | Maxim Zherelo | 0.68194 | 72 | Mais AbuSalah | 0.28201 |
| 23 | X4N7H055 | 0.75271 | 48 | Kolosov Arseny | 0.67784 | 73 | .... | 0.26981 |
| 24 | Hadjer | 0.75026 | 49 | Vlad Osipenko | 0.66425 | 74 | noreddine belhadj cheikh | 0.09071 |
| 25 | Salha Alzahrani | 0.73812 | 50 | drt1 | 0.66239 | | | |

The competition requests the top ranked teams to submit their codes at the end of the competition for final evaluation. Failure to do so results in the exclusion from the final evaluation. Six out of ten teams have submitted their codes. We verified their results of TEST1. Then, we did the final evaluation on TEST2. The reason for using TEST2 is to assess the models with a completely new set that was never seen by the model or the competitors. The performance of a model on TEST2 should be as good as on TEST 1.

Table 5 shows the results of TEST2 for the top-ranked teams. It can be observed that all the participating models outperformed the baseline results, as the baselines used very simple and basic methods. The top three best-performing teams were Wissam Antoun at first place with Macro-F1 = 0.79249. Both Abdullah I. Alharbi and Ali Salhi came in the second place since their Macro-F1 is about 0.79039. The CS-UM6P team came at third place with Macro-F1 = 0.78961.

Generally, all the top three ranked teams used MARBERT to get tweets representation. MARBERT [12] is a large-scale Arabic pre-trained model that focuses on both dialectal Arabic and modern standard Arabic. MARBERT is pre-trained over 6 Billions Arabic tweets. The teams followed similar steps for text pre-processing, such as:

- Removing unrecognizable symbols or characters that are not useful in understanding the text meaning; for instance, stop words, punctuation marks, diacritics, and elongations (tatweel).
- Replacing extra content in tweets with special tokens. For example "HASH" for hashtags, "USER" for user mentions, and "URL" for links.
- For emojis, the teams retained them as they are useful in emotion and sentiment analysis tasks. They also placed whitespace between any emojis to separate them and ensure they will be treated as separate words.

However, each team has its own approach for how to use MARBERT and other different deep learning techniques. Below, we provide a brief description of their approaches.

### 3.1 First Place Winner: Wissam Antoun Approach

This solution is based on an ensemble of 5 models with varying preprocessing and classifier design. All model variants are built over MARBERT. For classifier design, all models use a dense layer on top of MARBERT unless otherwise specified. Model training is done by hyperparameter grid-search with 5-fold cross-validation. Model I is a vanilla variant with only the general preprocessing steps applied. Model II enhances the emoji representation by replacing OOV emojis with ones that have a similar meaning.

Table 4: Other Metrics on official (Kaggle) results of the top 20 teams on TEST1 ranked by Macro-F1

| Team | Acc | Precision | Recall | Micro-F1 | Macro-F1 | Weighted-F1 |
| --- | --- | --- | --- | --- | --- | --- |
| GOF | 0.84945 | 0.80324 | 0.79656 | 0.84945 | 0.79985 | 0.84907 |
| Ahmed Elbehiry | 0.84940 | 0.80342 | 0.79630 | 0.84940 | 0.79979 | 0.84900 |
| Wissam Antoun | 0.84595 | 0.79868 | 0.79399 | 0.84595 | 0.79620 | 0.84581 |
| CS-UM6P | 0.84585 | 0.80053 | 0.78950 | 0.84585 | 0.79461 | 0.84539 |
| Ali Salhi | 0.84455 | 0.79848 | 0.79001 | 0.84455 | 0.79349 | 0.84442 |
| Taicheng Guo | 0.84280 | 0.79614 | 0.78625 | 0.84280 | 0.79097 | 0.84232 |
| Salma Jamal | 0.84165 | 0.79100 | 0.79093 | 0.84165 | 0.79090 | 0.84178 |
| Abdullah I. Alharbi | 0.83955 | 0.78802 | 0.79327 | 0.83955 | 0.79014 | 0.84033 |
| Aggies | 0.84060 | 0.79389 | 0.78237 | 0.84060 | 0.78753 | 0.84021 |
| KUIS AI | 0.84250 | 0.80158 | 0.77533 | 0.84250 | 0.78735 | 0.84105 |
| AraBrain Hidden Layers | 0.83820 | 0.78909 | 0.78123 | 0.83820 | 0.78482 | 0.83793 |
| AEM | 0.84000 | 0.80833 | 0.75453 | 0.83999 | 0.77837 | 0.83633 |
| Omar Mohamed | 0.82110 | 0.75697 | 0.79797 | 0.82110 | 0.77455 | 0.82481 |
| EAM | 0.83215 | 0.78413 | 0.76197 | 0.83215 | 0.77248 | 0.83041 |
| NLP players | 0.82425 | 0.76740 | 0.77468 | 0.82425 | 0.77074 | 0.82509 |
| raghad | 0.83080 | 0.78403 | 0.75604 | 0.83080 | 0.76910 | 0.82855 |
| Murtadha Aljubran | 0.82960 | 0.78083 | 0.75479 | 0.82960 | 0.76704 | 0.82757 |
| Roobaea Alroobaea | 0.82230 | 0.77044 | 0.75145 | 0.82230 | 0.76050 | 0.82089 |
| Yolo | 0.81815 | 0.75919 | 0.76122 | 0.81815 | 0.75998 | 0.81871 |
| Marwa Gharbi | 0.82950 | 0.79645 | 0.73114 | 0.82950 | 0.75917 | 0.82425 |

Table 5: Final evaluation for top-ranked teams on TEST2

| Team | Acc | Precision | Recall | Micro-F1 | Macro-F1 | Weighted-F1 |
| --- | --- | --- | --- | --- | --- | --- |
| Wissam Antoun | 0.84510 | 0.80669 | 0.78068 | 0.84510 | **0.79249** | 0.84382 |
| Abdullah I. Alharbi | 0.83975 | 0.79222 | 0.79017 | 0.83975 | **0.79039** | 0.84011 |
| Ali Salhi | 0.84301 | 0.80037 | 0.78158 | 0.84301 | **0.79039** | 0.84001 |
| CS-UM6P | 0.84520 | 0.81042 | 0.77198 | 0.84520 | **0.78961** | 0.84279 |
| Taicheng Guo | 0.81720 | 0.76040 | 0.75694 | 0.81720 | 0.75770 | 0.81743 |
| AraBrain Hidden Layers | 0.79275 | 0.78351 | 0.63320 | 0.79275 | 0.68273 | 0.77599 |

The participant noticed the repetitive use of "ورحمة الله وبركاته" and "السلام عليكم" in neutral tweets. This could confuse the classifier, if it encountered these words in for example a negative tweet, hence in Model III the variation of the phrase mentioned is removed before using fuzzy matching algorithms. In Model IV, the team tried to help the model by appending a sarcasm label to the input. They first trained a separate MARBERT on the ArSarcasm [14] dataset and then used it to label the training and test sets. Model V uses the vanilla preprocessing approach, but instead of a dense layer built on top of MARBERT, the team follows the approach detailed by Safaya et.al. [15] which uses a CNN-based classifier instead.

For the final prediction, the average prediction is computed from the predictions of the 5 models from cross-validation (this is done for each model separately). The participants noticed that the distribution of the predicted sentiment classes, doesn't quite match the true distribution, this is due to the model preferring the neutral class over the positive class. To counter that, they apply what they call Label-Weighted average, where after the final averaging they rescale each score of the three labels with a weight. The three weights were determined empirically.

### 3.2 Second Place Winner (I): Abdullah I. Alharbi Approach

The proposed method incorporates static character and word embeddings (CE and WE) and contextualized embeddings (MARBERT) to obtain a good representation for tweets. For static character and word embeddings, the team utilized ACWE model that was produced by [16]. ACWE combines static character- and word-level models to take advantage of each one of them. The word-level model was pre-trained on a large-scale dataset that consists of tweets written in a variety of Arabic dialects. It was constructed through the use of the word2vec algorithm as a means of learning how various individual words were represented. The character-level model is a pre-trained character representation model (CE) [17] that has been successfully employed in different Arabic affect tasks.

After ACWE model, CNN-LSTM architecture was employed as proposed in [18] to obtain the embeddings vectors after the training process. For generating contextual embeddings, the team also fine-tuned MARBERT model with the training data. Therefore, they concatenated both obtained vectors and fed it to a dense layer of number-of-classes output was put in place by leveraging softmax. Figure 2 illustrates the architecture of the proposed model.

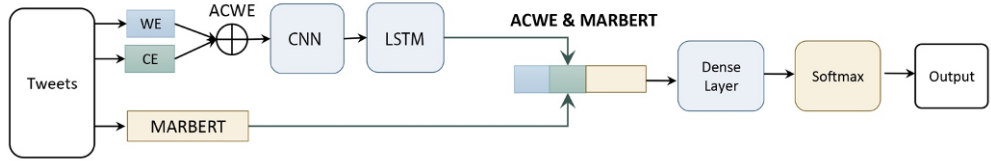

Figure 2: The architecture of Abdullah I. Alharbi Approach

### 3.3 Second Place Winner (II): Ali Salhi Approach

This solution is based on fine-tuning MARBERT with the training data by using different splitting ratios and training hyper parameters to get good performance. But the key factors in providing the high-performance model with steady results were the appending stems process and the feedback approach conducted on the training data.

Appending stems was implemented as an additional step for the data cleaning process. For each tweet, the stems of its words are extracted and added to it as extra text. The feedback approach was used to extend the training data by tweets from the test data. This was done by repeating the training with different N learning rates. Then, for a given test tweet, if the predicted labels in the N training trials are equals, the test tweet with its stable label will be added to the training data as an extra instance.

### 3.4 Third Place Winner: CS-UM6P Approach

The team proposed a deep multi-task model based on MARBERT encoder and task specific-attention layers [19, 20, 21]. In addition to the multi-class classification task, they have employed a one-versus-all (binary classification) task for each sentiment. To extract the task's discriminative features, they utilized four task attention layers on top of MARBERT's tokens embeddings. Each task's classifier is fed the output of its specific attention layer and the pooled output of MARBERT encoder. At the inference, the logits of binary tasks are concatenated and added to the logits of the multi-class classification to compute the probabilities of each polarity.

## 4 Related Work

Shared tasks allow the research community to work simultaneously towards solving a challenging problem. Shared tasks specify the main problem to be solved, and provide the research community with datasets, which are often challenging to obtain. Two similar shared tasks were organized in the past five years: SemEval 2017 - Task 4 [22], and WANLP 2021 - Subtask 2 [23]. Description of each shared task is provided next, and summary of the released datasets is provided in Table 6.

**SemEval 2017 - Task 4.** In 2017, SemEval announced a shared task which included Arabic dataset for the first time. Specifically, subtask A under SemEval-2017 Task 4 [22] released a dataset with a total of 9K Arabic tweets, divided into 35% for training and 65% for testing. Each tweet was labeled as either *positive, negative* or *neutral*. The distribution of positive/negative/neutral tweets in the training set is 22%/34%/44%, respectively, and 25%/36%39% in the testing set. The shared task has several subtasks, including overall sentiment analysis on a 2-point scale and a 3-point scale, as well as topic-level sentiment. A total of 8 teams participated in the Arabic subtask A, where the winning team used Naive Bayes classifier with a combination of lexical and sentiment features.

**WANLP 2021 - Subtask 2.** In 2021, a shared task was organized by the sixth Workshop on Arabic Natural Language Processing (WANLP) on sentiment detection in Arabic [23]. The shared task had two subtasks; sarcasam detection (subtask 1) and sentiment analysis (subtask 2). The dataset released for this shared task was ArcSarcasm-v2 [14], which has a total of 15K tweets divided into 80% training and 20% testing. The sentiment of each tweet was labelled as either *positive, negative* or *neutral*. The distribution of the positive/negative/neutral tweets in the training set is 17%/37%/46%, respectively, and 19%/56%/25% in the testing set. Additional labels were provided including sarcasm and dialect.

Table 6: Comparison of released datasets from similar tasks. Values indicate the number of tweets and percentage for each label in the training, testing and complete datasets, respectively.

|  | SemEval 2017 (Task 4) | | | WANLP 2021 (subtask 2) | | |
| --- | --- | --- | --- | --- | --- | --- |
|  | Training | Testing | Total | Training | Testing | Total |
| Positive | 743 (22%) | 1,514 (25%) | 2,257 (24%) | 2,180 (17%) | 575 (19%) | 2,755 (18%) |
| Negative | 1,142 (34%) | 2,364 (39%) | 3,506 (37%) | 4,621 (37%) | 1,677 (56%) | 6,298 (41%) |
| Neutral | 1,470 (44%) | 2,222 (36%) | 3,692 (39%) | 5,747 (46%) | 748 (25%) | 6,495 (41%) |
| Total | 3,355 (100%) | 6,100 (100%) | 9,455 (100%) | 12,548 (100%) | 3,000 (100%) | 15,548 (100%) |

Though the objective of both shared tasks is similar to KAUST's 2021 Arabic sentiment analysis competition, the size of the datasets released varies significantly. The dataset released in our competition, i.e., ASAD [11], is almost six times larger than ArcSarcasm-v2 [14] and ten times larger than SemEval 2017 Arabic sentiment analysis dataset [22].

## 5 Conclusion

We have described the data, evaluation process, and the results of "Arabic Sentiment Analysis Competition" organized by KAUST. The task is to analyze the sentiment of Arabic tweets by performing sentiment analysis classification. We received a high number of submissions, in total 1247 submission by 74 teams. Overall, the best systems used several preprocessing steps and they used the state-of-the-art pre-trained models such as MARBERT as well as different deep learning techniques. The final evaluation depends on the Macro-F1 results of a private testing file. The first-place winner has obtained 0.79249, and two participances shared the second place with Macro-F1=0.79039, and the third winner obtained 0.78961. We have made our annotated dataset ASAD publicly available to the research community beyond the competition. In addition to its main usage in the sentiment analysis, the dataset can be also used for other NLP tasks such as, dialect identification and spam detection. We will update the dataset with more annotations later.

## Acknowledgments

We would like to thank KAUST for sponsoring this competition. A great thank to the member of the host team Mohammad Al-Barazi. We also thank the winners: Wissam Antoun, Abdullah Alharbi, Ali Salhi, and Abdellah Elmekki for their kind help in describing their models. Finally, we thank Hassan Alzahrani, Arsalan Khatri, and Sarah Mitha for the IT support.